\documentclass[sigconf,authorversion,nonacm]{acmart}
\AtBeginDocument{%
  }

\usepackage{amsmath}
\usepackage{booktabs}       
\usepackage{graphicx}
\usepackage{caption}
\usepackage{subcaption}
\usepackage{float}
\usepackage{nicematrix}
\usepackage{enumitem}
\setlist[itemize]{leftmargin=*}
\usepackage{multirow}

\usepackage{natbib}
\begin{document}

\title[Synthetic and Manipulated Overhead Imagery]{Comprehensive Dataset of Synthetic and Manipulated Overhead Imagery for Development and Evaluation of Forensic Tools}


\author{Brandon B. May}
\affiliation{%
  \institution{STR}\city{Woburn}
  \state{MA}\country{USA}}
\email{brandonbmay@gmail.com}

\author{Kirill Trapeznikov}
\affiliation{%
  \institution{STR}\city{Woburn}
  \state{MA}\country{USA}}
\email{kirill.trapeznikov@str.us}

\author{Shengbang Fang}
\affiliation{%
  \institution{Drexel University}\city{Philadelphia}
  \state{PA}\country{USA}}
\email{sf683@drexel.edu}

\author{Matthew Stamm}
\affiliation{%
  \institution{Drexel University}\city{Philadelphia}
  \state{PA}\country{USA}}
\email{mcs382@drexel.edu}

\renewcommand{\shortauthors}{Brandon B. May, Kirill Trapeznikov, Shengbang Fang, \& Matthew Stamm}

\begin{abstract}
We present a first of its kind dataset of overhead imagery for development and evaluation of forensic tools. Our dataset consists of real, fully synthetic and partially manipulated overhead imagery generated from a custom diffusion model trained on two sets of different zoom levels and on two sources of pristine data. We developed our model to support controllable generation of multiple manipulation categories including fully synthetic imagery conditioned on real and generated base maps, and location. We also support partial in-painted imagery with same conditioning options and with several types of manipulated content. The data consist of raw images and ground truth annotations describing the manipulation parameters. We also report benchmark performance on several tasks supported by our dataset including detection of fully and partially manipulated imagery, manipulation localization and classification.
\end{abstract}

\begin{teaserfigure}
\centering
        \fbox{\includegraphics[width=.478 \textwidth]{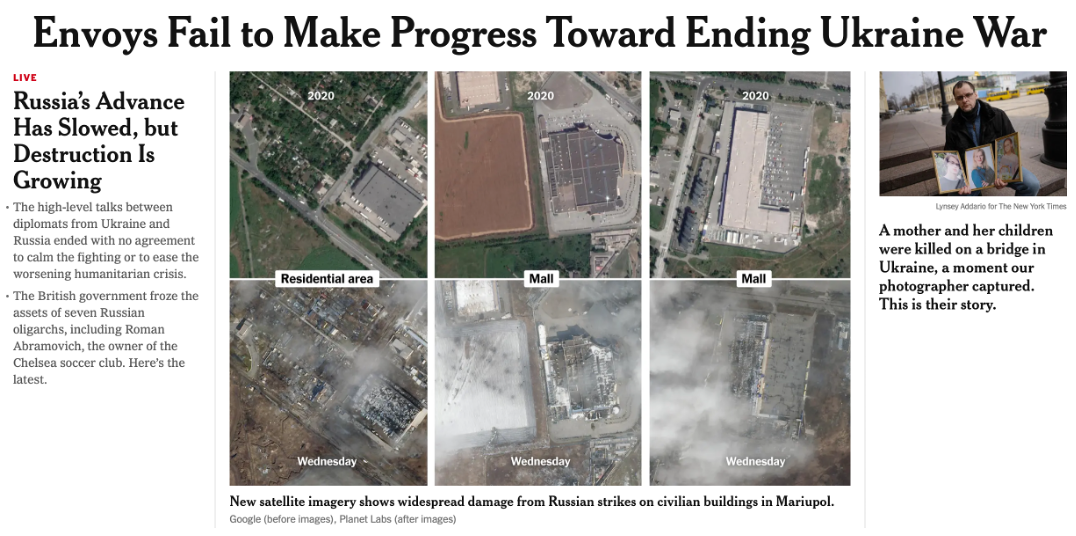}}
        \fbox{\includegraphics[width=.481 \textwidth]{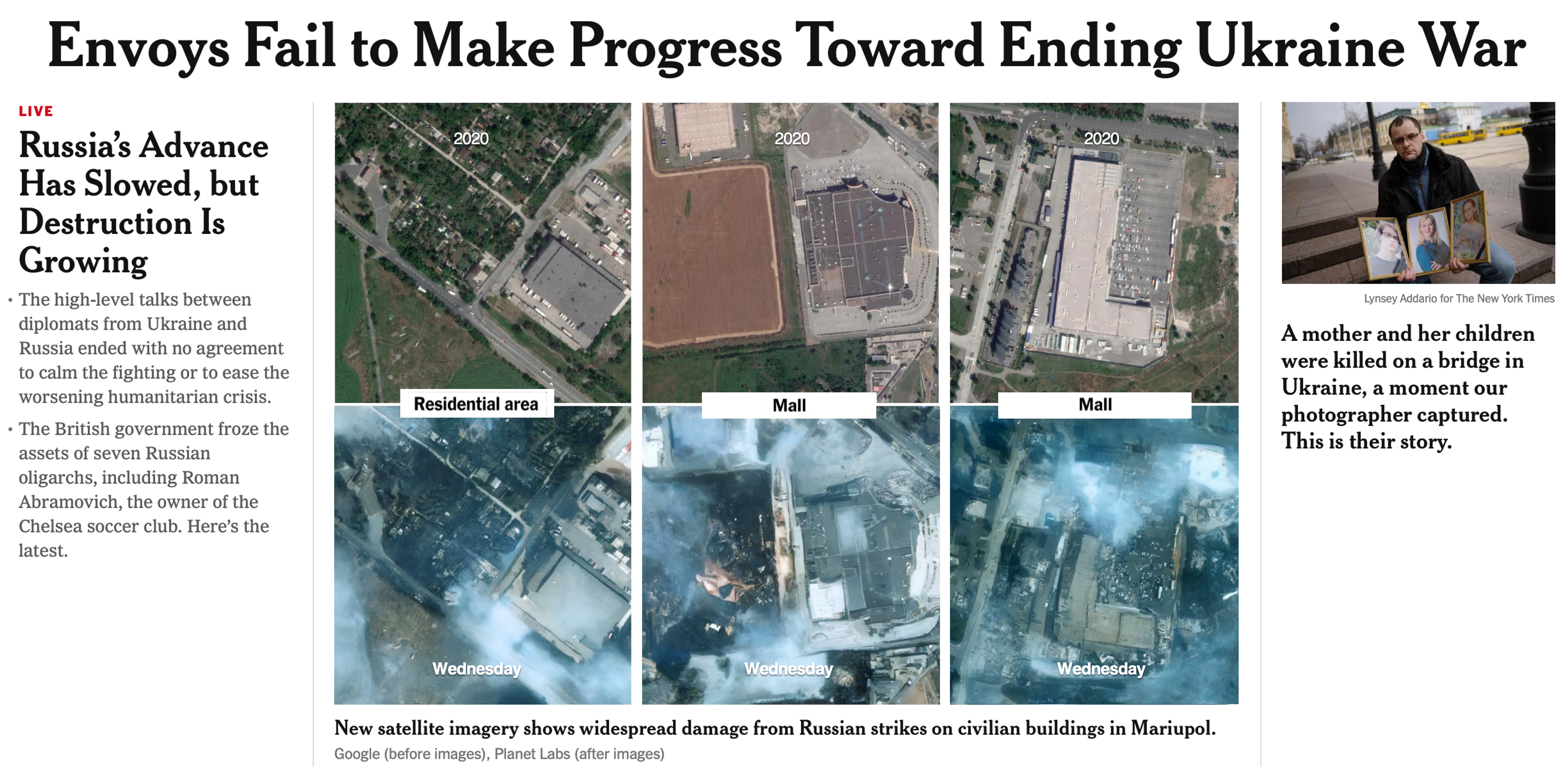}}
    \caption{Left: A New York Times homepage with before and after overhead views of bombings in Mariupol, Ukraine. Right: We highlight the threat of adversaries exploiting image generative techniques to manipulate information. We generate synthetic "aftermath" images (bottom row) by manipulating the original images using the diffusion models presented in this paper.}
    \label{fig:nytimes}
\end{teaserfigure}

\maketitle

\section{Introduction}


In the last several years, synthetic images have emerged as an important forgery threat. Various generative models, such as the StyleGAN-based family \cite{karras2020analyzing}, have made it possible to create realistic-looking images in various domains. As these models become more accessible to the general public, the potential for malicious uses of synthetic images increases. There have already been instances of StyleGAN-generated faces being used in influence campaigns \cite{unheard} and marketing scams \cite{linkedinai}.

The emergence of diffusion models \cite{dhariwal2021diffusion} poses a particularly important threat. With new conditional diffusion techniques (such as GLIDE \cite{nichol2022glide}, Imagen \cite{sahariaphotorealistic}, DALL-E 2 \cite{ramesh2022hierarchical} and Latent Diffusion \cite{rombach2022high}), users can control and modify specific aspects of the generation process through text, image or other inputs. With this capability, it is possible to selectively modify or synthesize particular parts of a real image. This enables the creation of partially synthetic images\footnote{In this paper, we'll use partially synthetic and partially manipulated interchangeably.}, i.e. images with both real and synthetic content, that are highly visually realistic. Existing research has focused largely on detecting GAN-generated content of mainly face images \cite{wang2022gan}, and nearly exclusively on discriminating between real and fully synthetic images, leaving an important gap in research.

Therefore, there is a need for datasets to advance research towards detecting synthetic images created using diffusion models.  The majority of existing synthetic datasets are composed of GAN-generated images \cite{verdoliva2020media}. However, because diffusion models likely leave behind different traces than GANs, new datasets are needed to train and benchmark new model detectors.  Additionally, existing datasets focus only on discriminating between fully real and fully synthetic images.  They do not consider partially synthetic images, such as those enabled by guided diffusion \cite{nichol2022glide}.  New datasets are needed to develop forensic systems capable of discriminating between these three types of images and localizing synthetic content. 

A parallel development has been the increasingly, commonplace use of satellite imagery by major news outlets as a way to provide context and supporting evidence for their reporting. This is partly due to wider access to satellite imagery from commercial companies. For instance in Fig. \ref{fig:nytimes} (left), the front page of the New York Times (from 16 March 2022) is using commercial satellite imagery to show a before and an after comparison of widespread damage from Russian strikes to civilian building in Mariupol, Ukraine. In addition to established news organizations, numerous social media accounts provide open source intelligence via analysis of publicly available overhead and ground level imagery. These accounts claim to analyze activities such as military buildup, construction, and the aftermath of natural disasters in locations all around the globe. (For examples see multiple investigations by bellingcat.com \cite{bellingcat})

We hypothesize that it's only a matter of time before synthetic forgeries of satellite imagery will be exploited for nefarious purposes such as disinformation campaigns (as we demonstrate in Fig. \ref{fig:nytimes} (Right)). So there is a clear need to develop state of the art forensics tools that can handle overhead imagery that has been manipulated with modern computer vision generative techniques. Specifically, the forensics need to be able to detect, localize and characterize manipulations in overhead imagery. While there has been active research in the field of digital and machine generated forensics, it has mostly focused on natural image domains, primarily due to the lack of high quality comprehensive datasets of overhead synthetically manipulated imagery.

\begin{figure}[tbh!]
    \centering
    \subcaptionbox{Pristine}{
        \includegraphics[angle=90,width=.23 \linewidth]{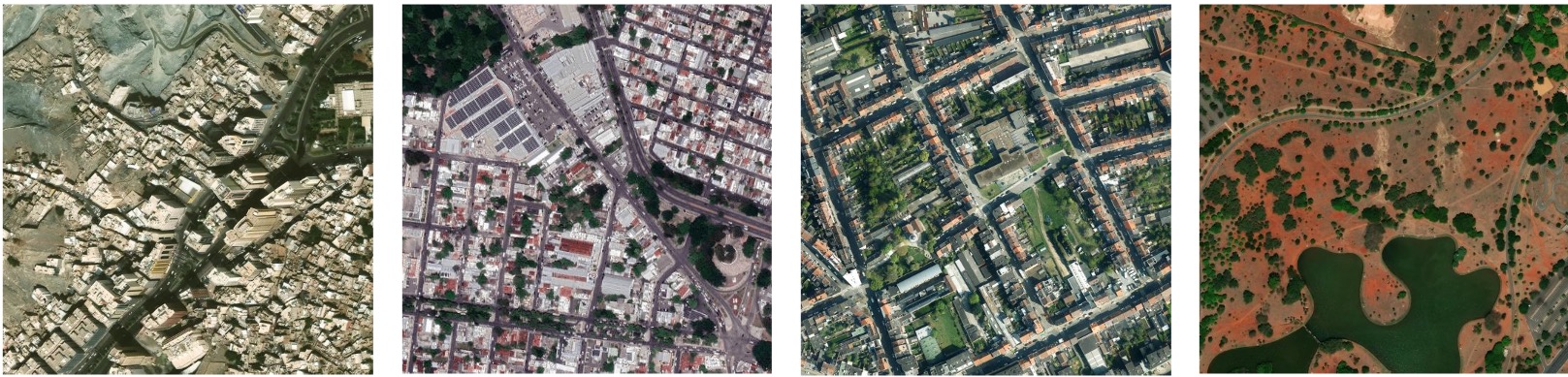}
    }
    \subcaptionbox{Synthetic}{
        \includegraphics[angle=90,width=.2268 \linewidth]{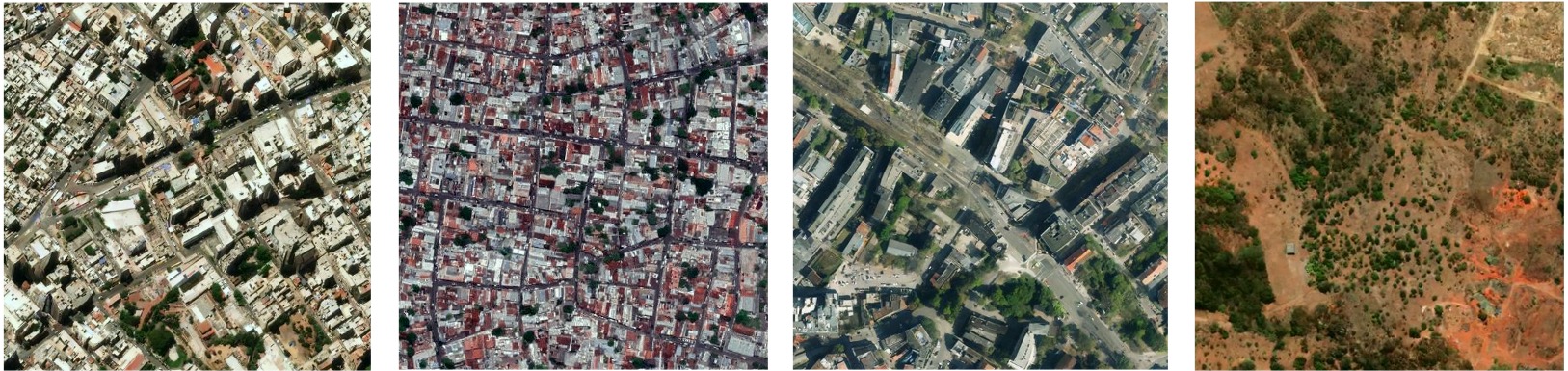}
    }
    \subcaptionbox{Partially Manipulated}{
    \includegraphics[angle=90,width=.4615 \linewidth]{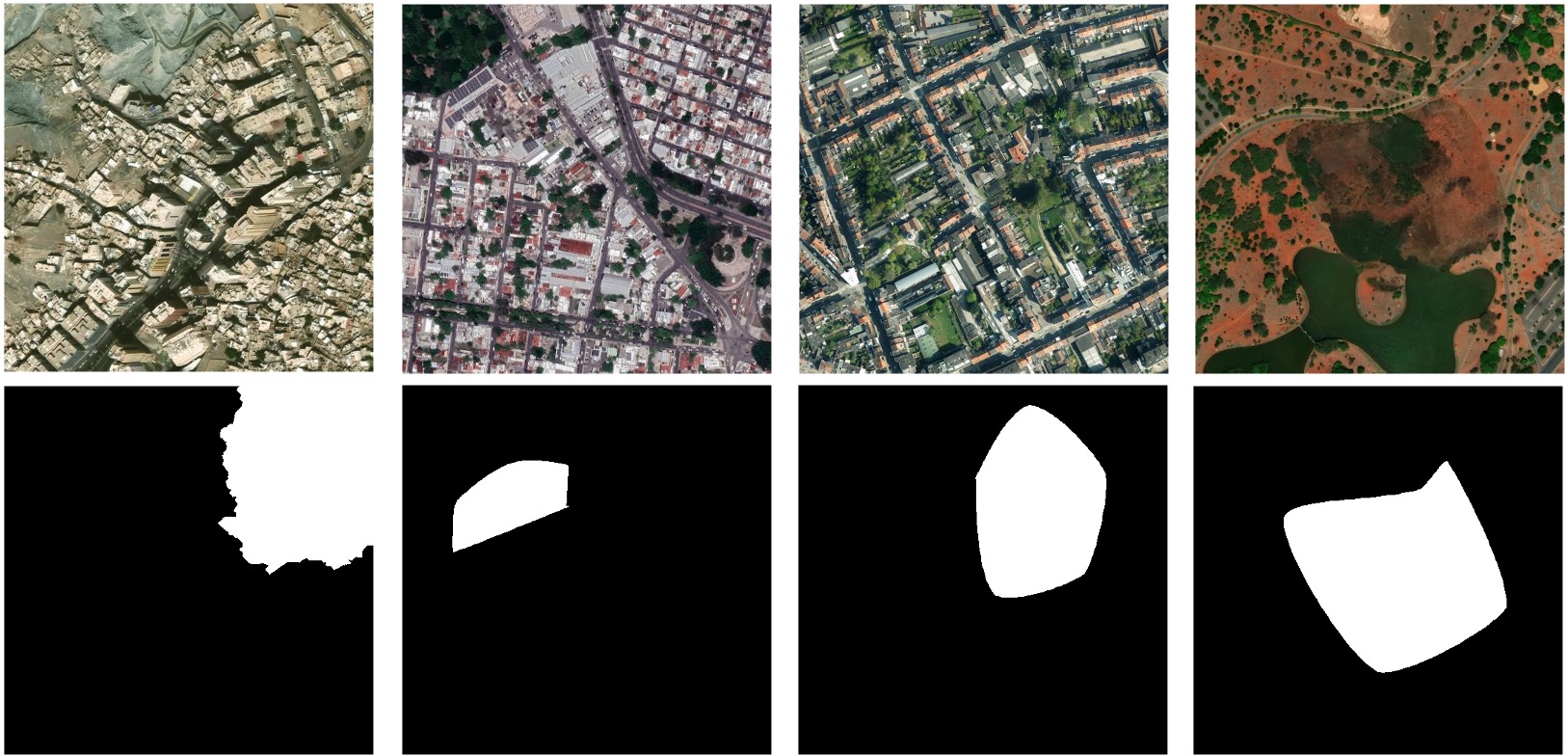}
    }
    \vspace{-3ex}
    \caption{Our dataset consist of three types of images (a) pristine, (b) fully synthetic, and (c) partially manipulated with ground truth manipulation masks and other parameters such as basemap type (none, truth, generated), city, image source (mapbox, google, zoom level), manipulation size and class (buildings-roads or greenspace-water). Rows corresponds to generation conditioned on a different city.}
    \label{fig:dataset_overview}
\end{figure}

In this paper, \textbf{(I)} we propose a new dataset of synthetic overhead imagery for forensic research created using diffusion models. Our dataset contains both real, fully and partially synthetic images along with localization masks (see Fig. \ref{fig:dataset_overview}). The full dataset can be accessed at \textbf{\texttt{\url{https://stresearch.github.io/synthetic-overhead-dataset/}}}. \textbf{(II)} The generation was done by our custom adaptation of guided diffusion \cite{nichol2022glide} trained on a large scale multi-source dataset of satellite imagery. Synthetic content was created using multiple strategies: unconditional, as well conditional generation from both real and \emph{synthetically generated} basemaps. Furthermore, we implemented models capable of mimicking natural disasters and performing partial image inpainting, which we used to create visually realistic, semantically meaningful forgeries (as we demonstrated with fake "aftermath" images in Fig. \ref{fig:nytimes}(right)).  
\textbf{(III)} Through a set of benchmarking experiments using several existing detection and localization algorithms, we show that important research is needed to improve performance on this dataset.

\section{Related Work}
To the best of our knowledge, we present the first application of diffusion models for generating overhead imagery. Diffusion models for image generation were initially popularized in a seminal paper by \cite{ho2020denoising}. Since then there have been many extensions focusing on unconditional, conditional and text-driven image generation and editing \cite{dhariwal2021diffusion,nichol2022glide,ramesh2022hierarchical,dhariwal2021diffusion,rombach2022high,sahariaphotorealistic}. However, the majority focuses on natural image domains such as human faces, animals and landscapes.

Specifically for overhead (satellite or aerial) imagery generation, existing techniques are based on generative adversarial networks (GANs) \cite{goodfellowgenerative}. There are several unconditional approaches based on StyleGAN-2 architectures such as \cite{this_city_does_not_exist,deepfake-satellite-images}. There are also several conditional generative approaches based on the image-to-image translation that can condition on additional information such as basemaps, elevation maps, etc \cite{rs13193984, Zhao_Deep, Zaytar_Satallite}. Our generative approach includes multiple additional conditioning options such as class and text guidance and it is based on diffusion models instead of GANs.

Other researchers have explored generative techniques for other modalities outside of visible spectrum such as SAR, infrared and hyper-spectral \cite{abady2020gan}. In this context, these approaches are designed to perform image translation between the modalities such as translating from visible to infrared. For a comprehensive review of remote sensing generative techniques (including conditional GAN-based visible light approaches) see \cite{abady2022overview}. 

To the best of our knowledge, there are no publicly available datasets that contain synthetic satellite imagery generated conditioned on multiple variables such as real and synthetic basemap, location and source providers. There are also no datasets that contain partially manipulated or in-painted imagery with a varying size of inpainted region and generated with same controls as above. \cite{deepfake-satellite-images} contains a dataset of real and StyleGAN2 generated synthetic images (trained from a single provider).

Prior work in forensic analysis for synthetic satellite imagery primarily focused on GANs \cite{abady2022overview}. There also has been recent work \cite{corvi2022detection,ricker2022towards,sha2022fake} exploring transferability of GAN trained synthetic image detectors to diffusion models for natural image domains. We are not aware of any work that benchmarks splicing or localization algorithms on diffusion model outputs.

\section{Dataset Description}

Our main contribution is a dataset for forensic research in synthetic overhead imagery. It consists of three major components: pristine, fully synthetic, and partially manipulated imagery (Fig. \ref{fig:dataset_overview}). The latter we found to be especially challenging for existing forensic methods. All imagery is either sourced directly from from MapBox and Google Maps satellite imagery or generated using our custom guided diffusion models trained with this data. Fully synthetic imagery for a given model and city combination was either generated unconditionally or conditioned on the corresponding truth or a synthetically generated basemap. Partially manipulated imagery for a given model and city combination was inpainted conditioned on an edited truth basemap. The editing was done by inpainting masked areas with buildings-roads or greenspace-water map layers. The breakdown of generation parameters is in Table \ref{tab:dataset_stats}. To prevent potential misuse, \emph{we are not releasing the full set of pristine imagery or the generative models themselves}.

\begin{table}[tb]

\begin{center}
\begin{NiceTabular}{p{1.8cm}|p{1.65cm}p{1.65cm}p{1.65cm}}
\midrule
& \textbf{Pristine} & \textbf{Synthetic} & \textbf{Manipulated} \tabularnewline
\midrule
\textbf{\# Images} & 4,964/1,511 & 2,496/886 & 2,540/753 \tabularnewline
\midrule
\textbf{Sources} & \Block{1-3}{MB16/MB16, G17, MB18; 152/80 cities} & & \tabularnewline
\midrule
\textbf{Basemaps} & \Block{1-3}{None, Truth, Generated, Inpainted} & & \tabularnewline
\midrule
\textbf{Manipulation} & \Block{1-3}{Buildings-Roads, Greenspace-Water} & & \tabularnewline
\midrule
\textbf{Masks} & \Block{1-3}{Bezier, GrabCut} & & \tabularnewline
\bottomrule
\end{NiceTabular}
\end{center}
\caption{\label{tab:dataset_stats} Dataset breakdown by each type of image (train/test)}

\end{table}

The dataset is partitioned into two splits: train and test for development and evaluation of forensic algorithms respectively. The set of cities covered in the splits are disjoint. For each of these splits, we selected at random a reference image to either be preserved as pristine or to be manipulated, with a further 50\% probability of being fully or partially synthetic. To test how well the performance of the forensic tools would generalize to unseen data sources and ground sampling distances, we generated all training data using the MB16 model and only included the G17 and MB18 models when generating the test data.

\subsection{Sourcing Pristine Imagery}
To train our generative models (see Sec. \ref{sec:diff_model}), we sourced a large collection of satellite image and basemap pairs. 
For each city and each image provider, we generated a collection of geospatial coordinates by uniformly sampling within the city bounds. These were used as the query locations when requesting image data. Specifically, we collected data from MapBox at zoom levels 16 (MB16) and 18 (MB18), and from Google Maps at zoom level 17 (G17 -- roughly equivalent to MB16) from 232 major non-US cities around the globe. The higher resolution MB18 imagery was only available and collected from 9 cities.

\begin{figure}[tb]
\captionsetup[subfigure]{labelformat=empty}
    \centering
    
    \subcaptionbox{}{
        \includegraphics[angle=-90, width=.27 \linewidth]{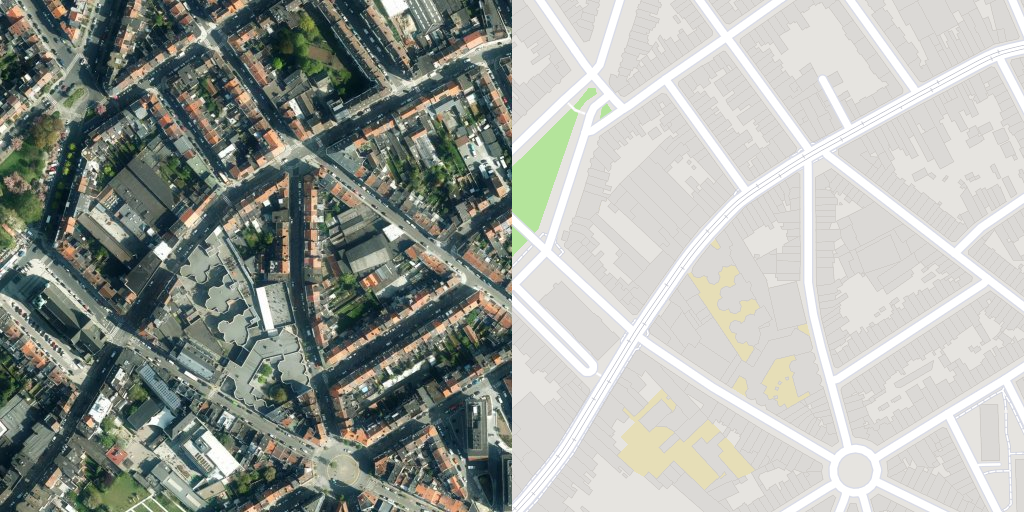}
    }
    \subcaptionbox{}{
        \includegraphics[angle=-90, width=.27 \linewidth]{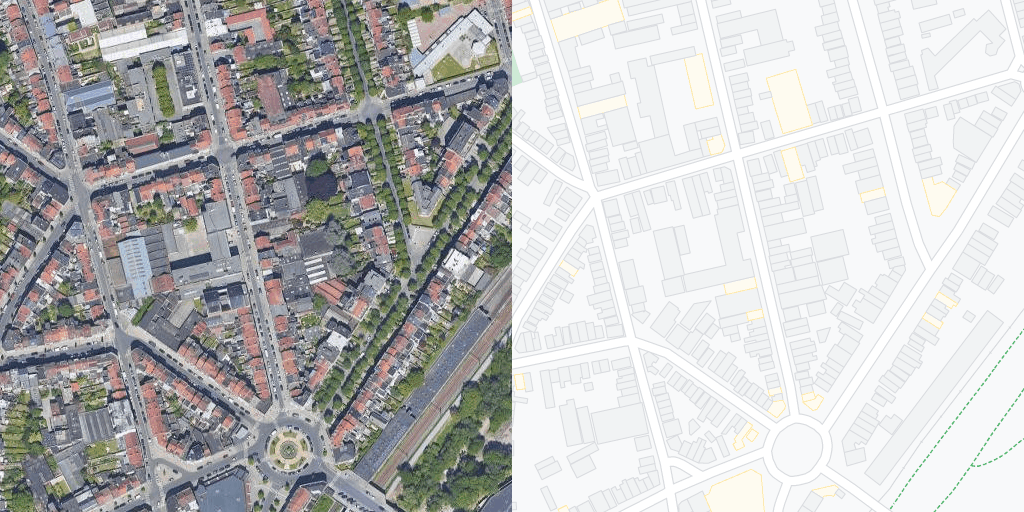}
    }
    \subcaptionbox{}{
        \includegraphics[angle=-90, width=.27 \linewidth]{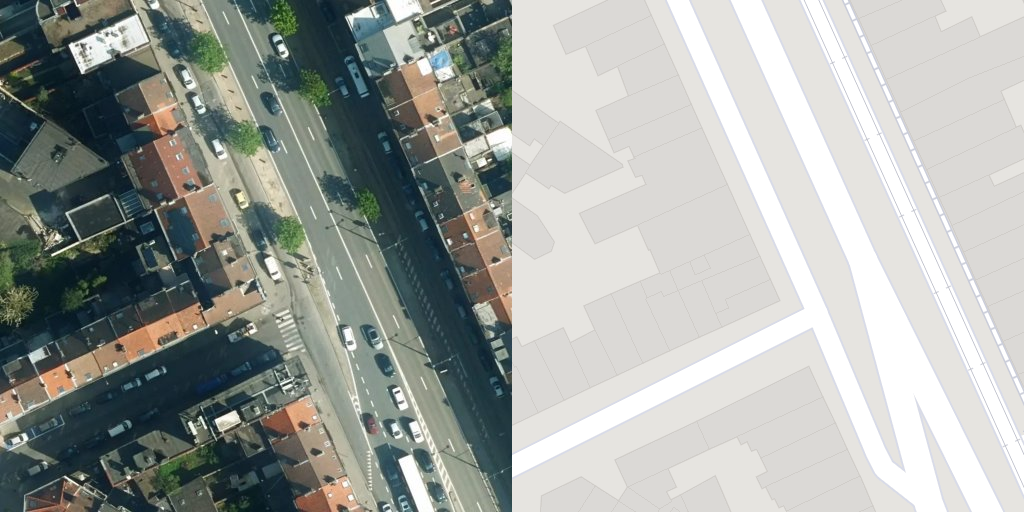}
    }
    \vspace{-7mm}
    \caption{All three varieties of data sources for satellite imagery and basemaps over approximately the same region in Brussels. From left to right: MapBox zoom level 16, Google Maps zoom level 17, MapBox zoom level 18.}
    \label{fig:sources_in_brussels}

\end{figure}

To simplify the basemaps for use as conditioning in our model, custom styles were created for each provider that removed extraneous elements such as borders, labels, and terrain shading leaving only color-coded representations of elements such as roads, highways, buildings, greenspace, water, and airports. Satellite and basemap image pairs have resolution of 512x512 centered on each coordinate for a given zoom level (16, 18 for MapBox, 17 for Google Maps). In total, we collected roughly 376k basemap-image pairs from MapBox and 77k basemap-image pairs from Google Maps. Fig. \ref{fig:sources_in_brussels} shows approximately the same location for all three data sources.

\subsection{Imagery with Synthetic Content}

\subsubsection{Fully Synthetic}
\begin{figure}[tb]
\captionsetup[subfigure]{labelformat=empty}
    \centering

    \subcaptionbox{}{
        \includegraphics[angle=-90,width=.27\linewidth]{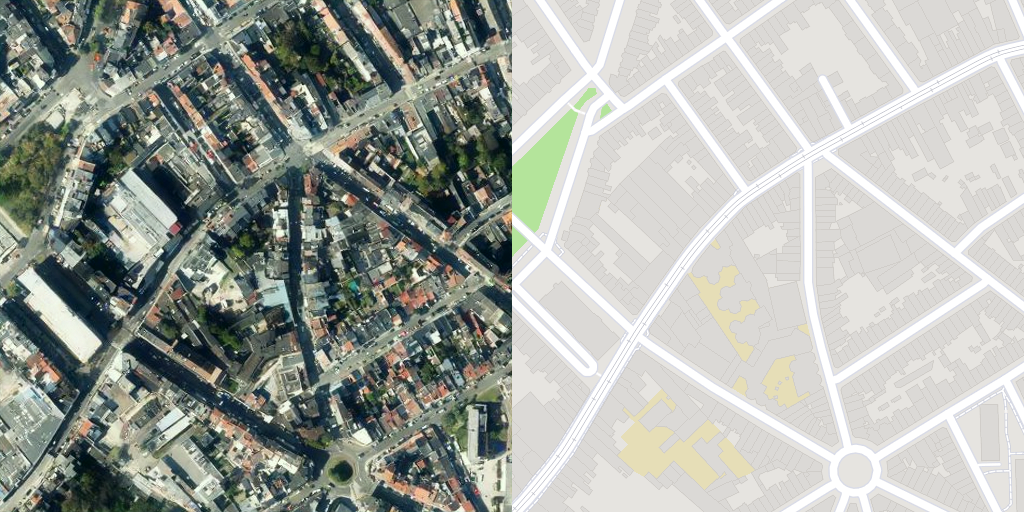}
    }
    \subcaptionbox{}{
        \includegraphics[angle=-90,width=.27\linewidth]{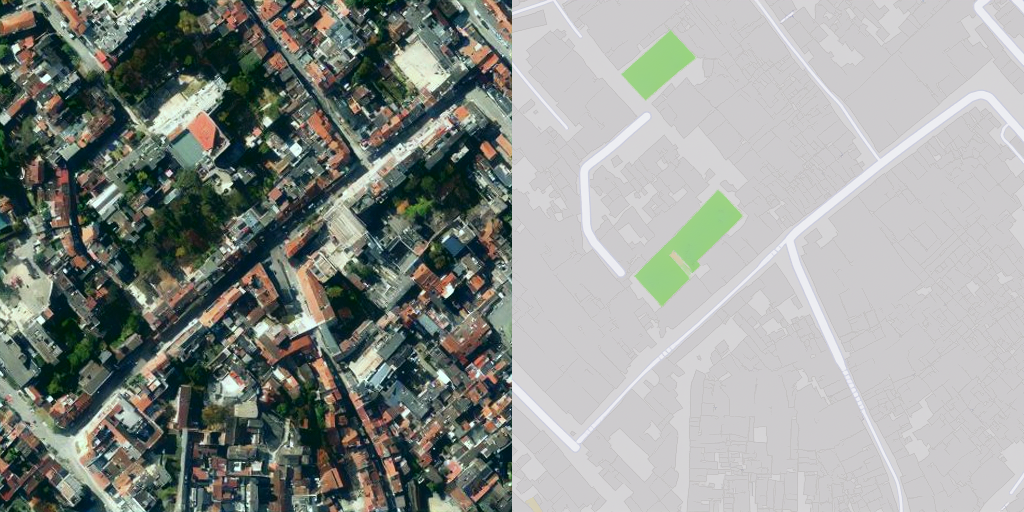}
    }
    \subcaptionbox{}{
        \includegraphics[angle=-90,width=.27\linewidth]{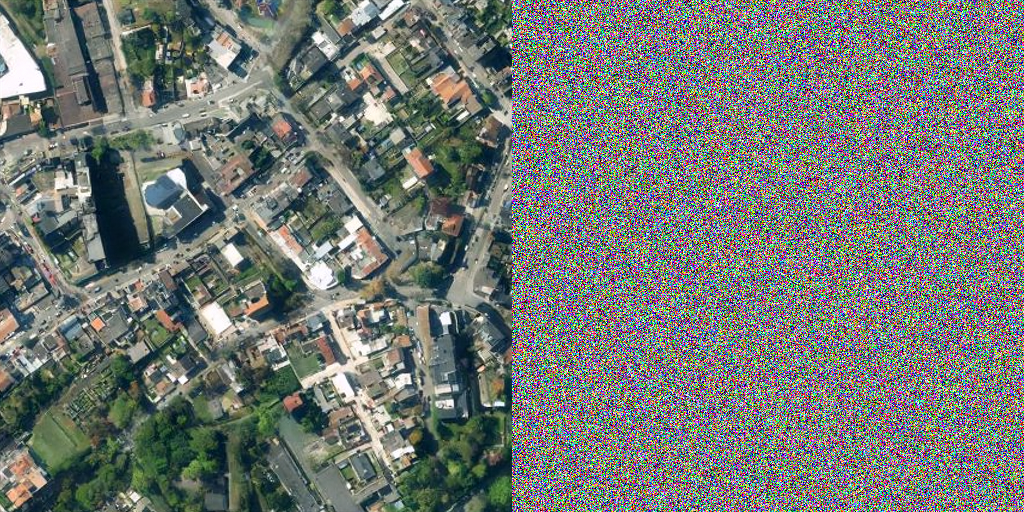}
    }
    \vspace{-5ex}
    \caption{Fully synthetic image generation examples for the city of Brussels. From left to right, conditioned on: truth basemap, generated basemap, no basemap.}
    \label{fig:fully_synthetic_brussels}

\end{figure}

Each fully synthetic image was generated using our basemap to satellite image diffusion model conditioned on the city and the associated randomly selected reference basemap. There are three basemap conditioning options; each occurring with equal probability in the dataset.
\textbf{(i) Truth:} the ground truth basemap was used as conditioning. \textbf{(ii) Generated:} the basemap was generated by using our basemap generation diffusion model, conditioned on the true city.
\textbf{(iii) None:} the basemap was generated by sampling from random Gaussian noise (See Fig. \ref{fig:fully_synthetic_brussels}).


\subsubsection{Partially Synthetic}

\begin{figure}[tb]
    \centering
    \includegraphics[angle=0,width=0.9\linewidth]{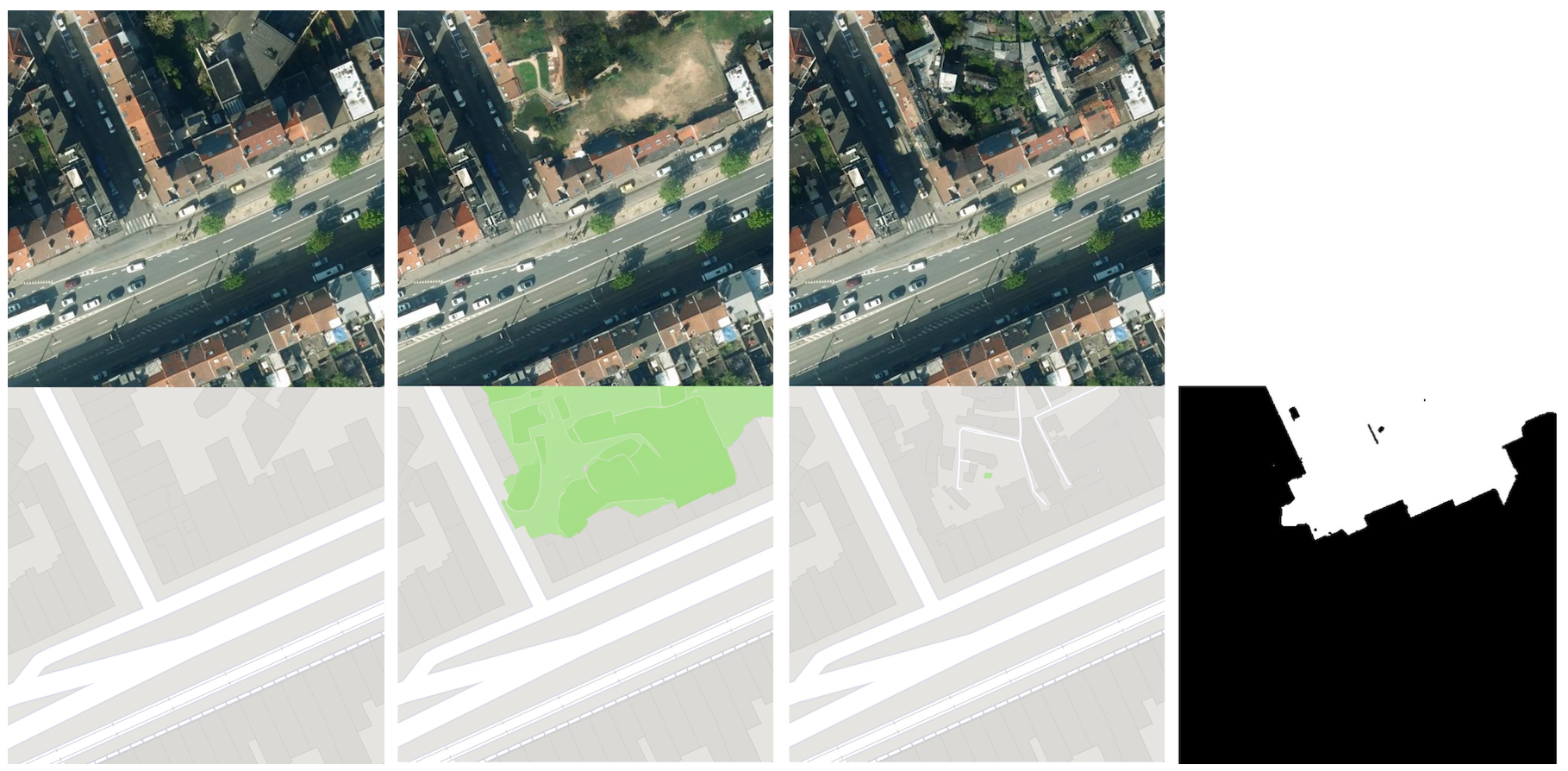}
    \vspace{-1ex}
    \caption{Partially manipulated generation example using MapBox zoom 18 high resolution imagery for the city of Brussels. Left to right: Ground truth, inpainted with greenspace-water, inpainted with buildings-roads, GrabCut mask.}
    \label{fig:partially_synthetic}

\end{figure}

One of the unique features of our dataset is to enable forensic research into detection and localization when images contain both real and synthetic content. This partially synthetic imagery were also generated using our basemap to satellite diffusion model, where masks were used to define the manipulated regions. These regions were then inpainted using a method similar to \cite{lugmayr2022repaint}, conditioned on one of two manipulation classes (buildings-roads or greenspace-water) and the city associated with the randomly selected reference image. The masks were randomly generated using either bezier shapes \cite{GitHubJviqueratbshapes} or GrabCut segmentations \cite{greig1989exact} across a range of sizes covering up to 20\% of the area of the image. In the case of the MB18 model, we used only GrabCut masks initialized with the building footprints from the ground truth basemap. Given the mask and manipulation class, the inpainting proceeded in a two step approach: first, inpainting of the ground \emph{truth basemap} to synthesize structure in the masked region according to the selected manipulation class, followed by inpainting of the satellite image conditioned on the manipulated basemap. Examples across models, manipulation classes, and mask types are in Fig. \ref{fig:partially_synthetic}.

\section{Guided Diffusion Approach} \label{sec:diff_model}

Our generative model used to create the dataset is based on an existing guided diffusion architecture \cite{nichol2022glide}. This presents additional challenges to forensics over GAN-based methods due to general lack of diffusion-trained detection models, better generation controls and ability to produce partially manipulated content. Additionally, we add support for the follow control mechanisms: (i) location conditioning using Classifier-Free Guidance (CFG), (ii)   basemap image-based conditioning, (iii) partial inpainting with basemap conditioning, (iv) compound editing, (v) CLIP \cite{radford2021learning} guidance and (vi) style transfer. Controls (i)-(iii) for generation and manipulation were used in constructing the dataset while (iv)-(vi) were not explicitly used but can still be potentially exploited for malicious uses and thus are presented as a proof of concept. For details on diffusion models and conditioning techniques consult \cite{ho2020denoising,nichol2022glide}.


\subsection{Controls and Guidance}

\subsubsection{Conditioning on Location}
To control the style of the generated image such that it matches a particular city in our dataset we apply class conditioning during training and inference. This is achieved by mapping the names of all cities to class IDs which are then used to construct an index of learned vector embeddings that get appended to the timestep embeddings at each residual block in the UNet architecture of the model. Using the embedding corresponding to a specific city at inference time, the model generates images in the desired style. See Fig. \ref{fig:dataset_overview}(b) for examples of fully synthetic images generated in the style of four different cities.
\subsubsection{Conditioning on Basemap} \label{sec:basemap_conditioning}
Support for conditioning on a reference basemap was added by expanding the number of input channels from three to six to support an additional RGB image as input to the network. During training the basemap associated with the overhead image is passed along as the conditioning image. At inference time, if a basemap is supplied, the generated image will have structures very closely corresponding to those specified by the basemap. Note that this technique can be used for training any kind of image-to-image translation model. The left and center images in Fig. \ref{fig:fully_synthetic_brussels} show examples of fully synthetic outputs generated conditioned on an input basemap.
\subsubsection{Partial Inpainting Support}
We implemented an inpainting technique closely following the approach in \cite{lugmayr2022repaint} that was used to perform partial manipulation of the images. Given a mask specifying the pixels to be generated (and thus also the pixels that are ``known"), at each iteration of the denoising process we replace the region of ``known" pixels in the generated image with the ground truth data according to the mask, noised using the learned denoising function up to the appropriate timestep. This effectively conditions each subsequent denoising step on the ground truth surrounding the inpainted region ensuring a coherent result. Figs. \ref{fig:dataset_overview}(c) and \ref{fig:partially_synthetic} both show examples of partial manipulations using this inpainting technique. The white regions of the masks represent the generated pixels and the black regions represent the ``known" pixels in the corresponding basemap and overhead images.

\subsubsection{Compound Editing}
\begin{figure}[tb]
\captionsetup[subfigure]{labelformat=empty}
    \centering

    \subcaptionbox{}{
        \includegraphics[angle=-90,width=.22\linewidth]{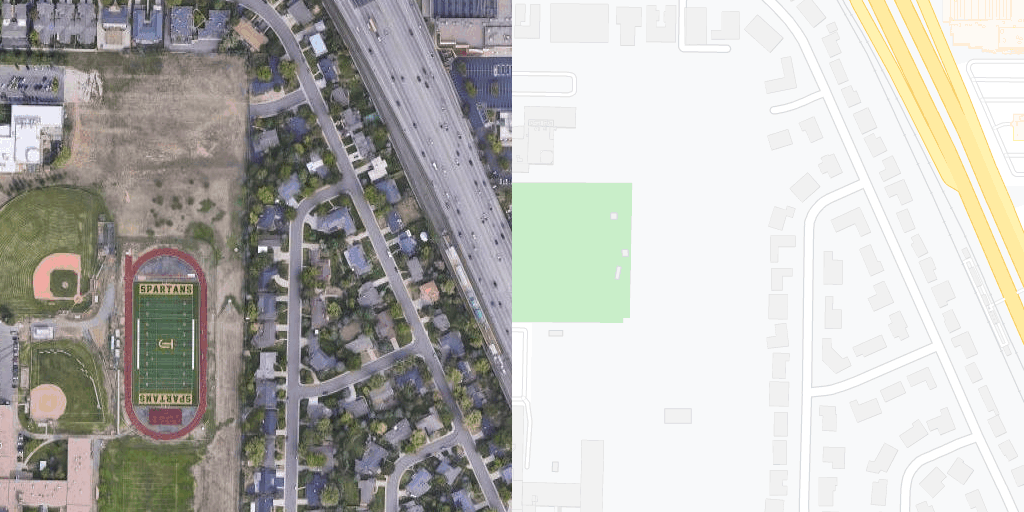}
    }
    \subcaptionbox{}{
        \includegraphics[angle=-90,width=.22\linewidth]{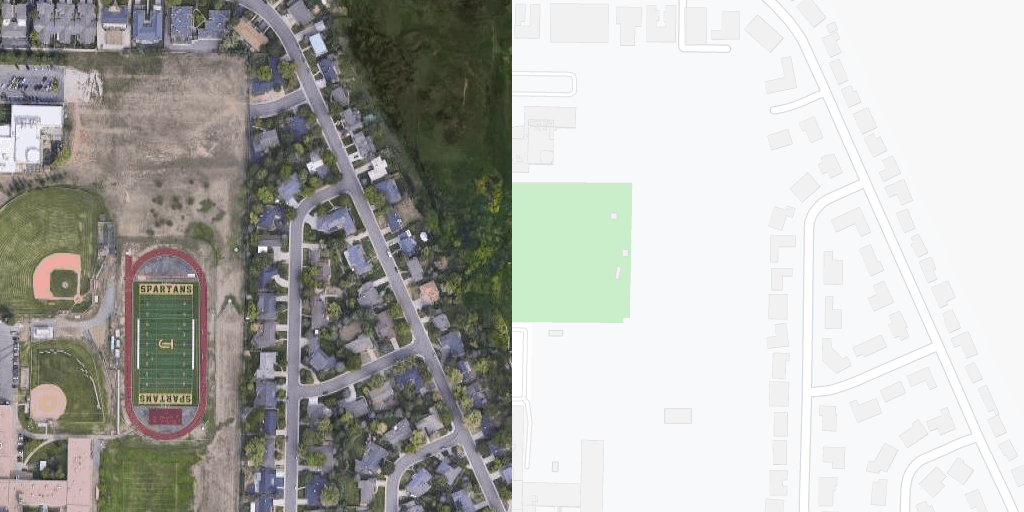}
    }
    \subcaptionbox{}{
        \includegraphics[angle=-90,width=.22\linewidth]{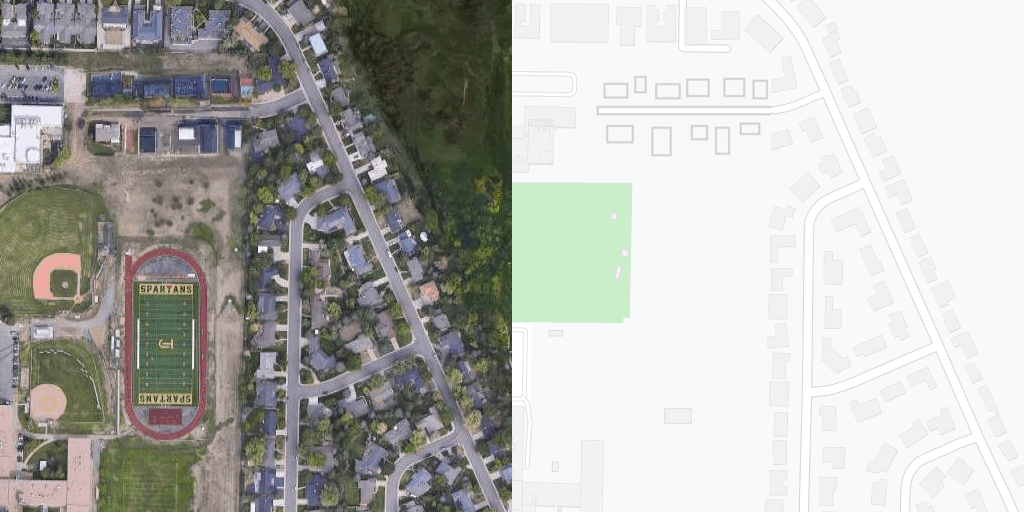}
    }
    \subcaptionbox{}{
        \includegraphics[angle=-90,width=.22\linewidth]{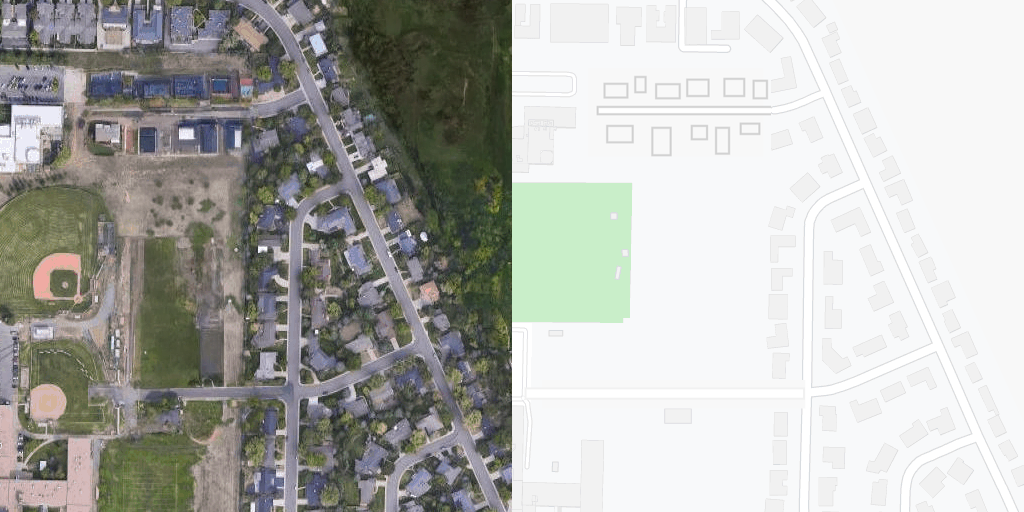}
    }
    \vspace{-5ex}
    \caption{Compound editing by manual basemap manipulation. Left to right: ground truth, highway and business park deleted and replaced with forest, road on right extended and populated with buildings, sports field erased and added connecting road going through neighborhood.}
    \label{fig:compound_editing}

\end{figure}

Partial manipulations can also be used to perform ``compound" editing of a reference image by iteratively manipulating its basemap, manually or otherwise. The output of each previous edit is used as the input to the next stage until the desired manipulations are complete. For example, Fig. \ref{fig:compound_editing} demonstrates the results of manually editing the basemap several times to replace highways with forest and add in new roads and buildings. At each stage the masks are computed automatically based on the edited regions of the basemap, and the corresponding regions of the image are inpainted conditioned on the new basemap.

\subsubsection{CLIP Guidance}
Trained diffusion models can generate content conditioned on external information provided by other models such as image classifiers. During inference, at each denoising step, the gradient of the classifier score w.r.t. to the generated image modifies the output by a scalar guidance factor \cite{nichol2022glide}. We use the CLIP multi-modal model (fine-tuned on satellite imagery) to guide the content of images using text prompts without having to explicitly train the diffusion model with text-image pairs.
\subsubsection{Style Transfer of Natural Disasters}
\begin{figure}[tb]
\captionsetup[subfigure]{labelformat=empty}
    \centering

    \subcaptionbox{}{
        \includegraphics[width=.3\linewidth]{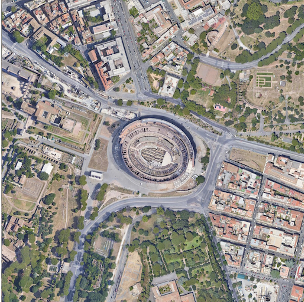}
        \vspace{-3ex}
    }
    \subcaptionbox{}{
        \includegraphics[width=.3\linewidth]{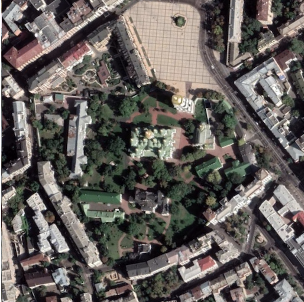}
        \vspace{-3ex}
    }
    \subcaptionbox{}{
        \includegraphics[width=.3\linewidth]{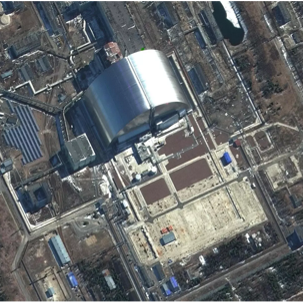}
        \vspace{-3ex}
    }
    \subcaptionbox{}{
        \includegraphics[width=.3\linewidth]{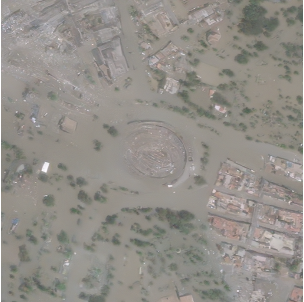}
    }
    \subcaptionbox{}{
        \includegraphics[width=.3\linewidth]{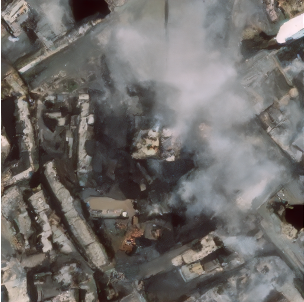}
    }
    \subcaptionbox{}{
        \includegraphics[width=.3\linewidth]{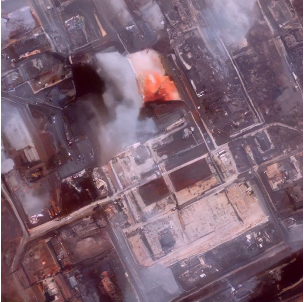}
    }
    \vspace{-5ex}
    \caption{Generating natural disasters using style transfer. Top row: original images. Bottom row: synthetic disaster images. Left to right: Colosseum (Rome) flooded with `Hurricane Harvey" style, St. Sophia's Cathedral (Kiev) and Nuclear Sarcophagus (Chernobyl). ``SoCal Fire" style and text prompt ``destroyed buildings with smoke rising."}
    \label{fig:disasters}
\end{figure}

The image-to-image translation method of \S\ref{sec:basemap_conditioning} can be used to train a ``style-transfer" model given pairs of images from the source and target distributions. The model learns to generate an image from the target distribution conditioned on a reference image from the source. We train a natural disaster generator using the xView2 dataset \cite{gupta2019creating} conditioned on the disaster class and the source imagery. The xView2 dataset contains before and after images from around the world of a variety of natural disasters including hurricanes, wildfires, tsunamis, and volcanoes. Fig. \ref{fig:disasters} shows examples of how we can construct realistic disaster imagery (bottom row) of particular locations using the original images (top row), natural disaster class conditioning, and prompt conditioning via CLIP guidance.

\subsection{Implementation and Training}
Each diffusion model was trained from scratch across 4 NVIDIA GTX 2080 GPUs for about 100k iterations with a batch size of 16, taking approximately one week. Compared to the default model parameters used by guided diffusion, we increased the output resolution from 256x256 to 512x512 and decreased the number of channels from 256 to 128 to fit within the memory constraints. Otherwise, we used the same default parameters corresponding to the 256x256 (unconditional) pretrained ImageNet model \cite{nichol2022glide}. Simple random vertical and horizontal flips were used for our data augmentations. During inference, denoising was performed for approximately 1000 time steps. The final output was then color matched to the reference image to better preserve realistic dynamic range.

\section{Baseline Forensic Performance}

We evaluate several existing forensic algorithms on the following tasks: (1) detection of fully synthetic imagery, (2) detection of partially synthetic imagery, (3) detection of any synthetic content (full or partial) in image and (4) localization of manipulated region. 

We employ the following existing forensic models as baselines. \textbf{Synthetic Image Detection:} GAN detector from \cite{wang2020cnn} trained to detect images generated from several CNN-based ProGAN model and shown to have good generalization to other generators. \textbf{Partial or Splicing Detection and Localization: } We evaluate three well established models, Forensic Similarity Graphs (FSG) \cite{mayer2020exposing}, EXIFNet \cite{huh2018fighting} and Noiseprint \cite{cozzolino2019noiseprint} (for localization only). \emph{We used the models as is} i.e. trained on their respective datasets and did not further fine-tune them on our proposed dataset.

\subsection{Detection of Synthetic Images}

Classification of fully synthetic vs pristine images is a well established task in forensic research, and we evaluate the GAN detector described previously. However, there are no existing classifiers for discriminating between pristine, fully synthetic and partially synthetic. For real world forensic application, differentiating between fully synthetic and partially manipulated (or spliced) images is critical. In order to be able to localize and characterize manipulations, first, one must determine if only part of the image has been manipulated and only then apply localization algorithms. Furthermore (as we discover), existing synthetic image detectors are not always well suited to classify images that are not fully synthetic.

\newcommand{\NA}{\multicolumn{1}{c}{---}}
\begin{table}[tb]
    \setlength{\tabcolsep}{1pt}
    \begin{tabular}{m{30mm}m{18mm}m{18mm}m{18mm}}
        \toprule
        \multicolumn{1}{c}{\multirow{2}{*}{\textbf{\shortstack{Benchmark\\Task}}}} & \multirow{2}{*}{\shortstack{\textbf{GAN \cite{wang2020cnn}}\\\textbf{Detector }}} & \multicolumn{2}{c}{\textbf{Splicing Detector}} \\
        & & \multicolumn{1}{c}{FSG \cite{mayer2020exposing}} & \multicolumn{1}{c}{EXIF \cite{huh2018fighting}} \\
        \midrule
        Pristine v.s. Fully & \textcolor{blue}{0.82}/\textcolor{teal}{0.75}/\textcolor{red}{0.50}  & \NA & \NA \\
        Pristine v.s. Partially & \textcolor{blue}{0.57}/\textcolor{teal}{0.56}/\textcolor{red}{0.50} & \textcolor{blue}{0.62}/\textcolor{teal}{0.59}/\textcolor{red}{0.50}  & \textcolor{blue}{0.54}/\textcolor{teal}{0.54}/\textcolor{red}{0.50} \\
        Pristine v.s. Any &  \textcolor{blue}{0.69}/\textcolor{teal}{0.64}/\textcolor{red}{0.50} & \NA & \NA \\
        Partially v.s. Fully & \NA & \textcolor{blue}{0.60}/\textcolor{teal}{0.58}/\textcolor{red}{0.50} & \textcolor{blue}{0.59}/\textcolor{teal}{0.57}/\textcolor{red}{0.50} \\
        \bottomrule
    \end{tabular}
    \caption{Results on four binary tasks. Metrics are \textcolor{blue}{AUC} / \textcolor{teal}{Accuracy w. threshold calibration} / \textcolor{red}{Accuracy w. original threshold}}.
    \label{tab:detection}
\end{table}

We map the problem of reasoning over varying amount of synthetic content to four binary tasks: (i) pristine vs fully synthetic, (ii) pristine vs partially synthetic, (iii) pristine vs any synthetic and (iv) partially vs fully synthetic. The results are presented in Table~\ref{tab:detection}. For the GAN-synthetic detector, we assess its ability to discriminate tasks (i) - (iii). We report the AUC of the ROC for detection and also the average accuracy before and after calibration on our proposed dataset. The accuracy before calibration uses the threshold proposed in the original work, while the accuracy after calibration uses the threshold that maximizes the average accuracy. The average accuracy before calibration is 0.5, as the original threshold causes all algorithms to always predict the image to be non-synthetic or non-spliced. For the GAN detector, it performs well in discriminating between real and fully synthetic images, with an AUC of 0.82. However, it only has an AUC of 0.57 in detecting partially synthetic images. This indicates that the performance drops sharply when only part of the image is synthetic. Similar drop in non-calibrated performance has been observed in prior work \cite{corvi2022detection} when testing GAN-trained detectors on diffusion model outputs.

For the partial manipulation (or splicing) detection, we evaluate FSG and EXIFNet performance in discriminating pristine vs partially synthetic, and partially synthetic vs fully synthetic. Both detectors struggle in these tasks. This aligns with our expectation, as they are not trained on our proposed synthetic overhead images, which may have different forensic traces from the camera model images that they were trained on.

\subsection{Pristine vs. Fully vs. Partially Synthetic}

To construct a 3-way classification that can discriminate pristine, partial, and fully synthetic images, we combine the output decisions of the GAN detector with a splicing detector. We choose two decision strategies. In the first one, the GAN detector comes first. If it classifies an input as pristine then the image is predicted to be pristine. Otherwise, we use the splicing detector to classify partial (spliced) or fully (non-spliced) synthetic. In the second strategy, the splicing detector comes first. If the splicing detector finds a splice then the image is classified to be partially synthetic. Otherwise, we use the GAN detector to detect if it is fully synthetic or pristine.

For evaluation, we test the performance of the combinations of (FSG, GAN Detector) and (EXIFNet, GAN Detector), and report the best average accuracy over three classes. Noting that the average accuracy would be 0.33 for a random guess. We report that the two combinations of the first hierarchical system have a lower average accuracy of 0.36 (with FSG) and 0.43 (with EXIFNet), while the combinations of the second hierarchical system have similar performance, with an average accuracy of 0.51 but still not useful for real world applications.



\subsection{Localizing Manipulated Regions}
If the the image is deemed to be partially manipulated then the next step in the processing chain is to localize the manipulated area. We evaluate the performance of existing forgery localization algorithms including FSG, EXIF-Net and Noiseprint on the  partially synthetic images. The evaluations are divided into 4 categories based on the different sizes of synthetic content in the images, and the Matthews Correlation Coefficient (MCC) score is calculated using the predicted and ground truth masks. The results in Table \ref{tab:baseline_results} show that FSG has the best performance on large synthetic content, with an MCC score of 0.350. Although FSG and EXIF-Net achieved an MCC of 0.82 and 0.78 on the Columbia splicing dataset \cite{ng2004data}, their performance drops significantly on our dataset. Note that we observed similar performance drops between calibrated and uncalibrated scores in the detection tasks. Further investigation and research are necessary to address this issue.

\begin{table}[tb]
\begin{center}
\begin{tabular}{lrrrcrrr}
\toprule
& \multicolumn{7}{c}{\textbf{Baseline Methods}}\\
& \multicolumn{3}{c}{Calibrated} && \multicolumn{3}{c}{Uncalibrated}\\
\cmidrule{2-4} \cmidrule{6-8}
\textbf{Size} & EXIF & FSG & NP & \phantom{a} & EXIF & FSG & NP\\
\midrule
Large & 0.194 & \textbf{0.350} & 0.325 & & 0.063 & \textbf{0.348} & 0.067\\
Medium & 0.150 & 0.327 & \textbf{0.331} & & 0.055 & \textbf{0.326} & 0.072\\
Small & 0.115 & \textbf{0.269} & 0.250 & & 0.028 & \textbf{0.265} & 0.070\\
X-Small & 0.055 & 0.143 & \textbf{0.163} & & 0.014 & \textbf{0.136} & 0.058\\
\midrule
\textbf{Overall} & 0.133 & \textbf{0.279} & 0.274 & & 0.043 & \textbf{0.277} & 0.068\\
\bottomrule
\end{tabular}
\end{center}
\caption{Localization using baselines, by region size, measured by MCC. Calibrated results computed using best performing mask threshold per method on our dataset. Uncalibrated results computed on the Carvalho DSO-1 dataset \cite{Carvalho_dso1}.}
\label{tab:baseline_results}
\end{table}

\section{Conclusion}
In this paper, we highlight the need for further research into detection and localization of synthetic content in overhead imagery generated by diffusion models. Specifically, we release a first of its kind dataset of real, fully and partially synthetic imagery. The data is generated with a custom implementation of a guided diffusion model with support for multiple manipulation methods. Lastly, we benchmark several baseline forensic models to illustrate the importance of continuing forensic research in the space of diffusion models and partially synthetic content localization.


\section{Acknowledgements}
This material is based on research sponsored by DARPA and AFRL 
(HR0011-20-C-0126, HR0011-20-C-0129).


\bibliography{arxiv_references}

\begin{thebibliography}{34}
\providecommand{\natexlab}[1]{#1}
\providecommand{\url}[1]{\texttt{#1}}
\expandafter\ifx\csname urlstyle\endcsname\relax
  \providecommand{\doi}[1]{doi: #1}\else
  \providecommand{\doi}{doi: \begingroup \urlstyle{rm}\Url}\fi

\bibitem[Karras et~al.(2020)Karras, Laine, Aittala, Hellsten, Lehtinen, and
  Aila]{karras2020analyzing}
Tero Karras, Samuli Laine, Miika Aittala, Janne Hellsten, Jaakko Lehtinen, and
  Timo Aila.
\newblock Analyzing and improving the image quality of stylegan.
\newblock In \emph{CVPR}, pages 8110--8119, 2020.

\bibitem[unh()]{unheard}
Graphika and stanford internet observatory (2022). unheard voice: Evaluating
  five years of pro-western covert influence operations.
\newblock URL \url{https://purl.stanford.edu/nj914nx9540}.

\bibitem[Bond()]{linkedinai}
Shannon Bond.
\newblock That smiling linkedin profile face might be a computer-generated
  fake.
\newblock URL
  \url{https://www.npr.org/2022/03/27/1088140809/fake-linkedin-profiles}.

\bibitem[Dhariwal and Nichol(2021)]{dhariwal2021diffusion}
Prafulla Dhariwal and Alexander Nichol.
\newblock Diffusion models beat gans on image synthesis.
\newblock \emph{Advances in Neural Information Processing Systems},
  34:\penalty0 8780--8794, 2021.

\bibitem[Nichol et~al.(2022)Nichol, Dhariwal, Ramesh, Shyam, Mishkin, Mcgrew,
  Sutskever, and Chen]{nichol2022glide}
Alexander~Quinn Nichol, Prafulla Dhariwal, Aditya Ramesh, Pranav Shyam, Pamela
  Mishkin, Bob Mcgrew, Ilya Sutskever, and Mark Chen.
\newblock Glide: Towards photorealistic image generation and editing with
  text-guided diffusion models.
\newblock In \emph{ICML}, pages 16784--16804. PMLR, 2022.

\bibitem[Saharia et~al.()Saharia, Chan, Saxena, Li, Whang, Denton, Ghasemipour,
  Gontijo-Lopes, Ayan, Salimans, et~al.]{sahariaphotorealistic}
Chitwan Saharia, William Chan, Saurabh Saxena, Lala Li, Jay Whang, Emily
  Denton, Seyed Kamyar~Seyed Ghasemipour, Raphael Gontijo-Lopes, Burcu~Karagol
  Ayan, Tim Salimans, et~al.
\newblock Photorealistic text-to-image diffusion models with deep language
  understanding.
\newblock In \emph{Advances in Neural Information Processing Systems}.

\bibitem[Ramesh et~al.(2022)Ramesh, Dhariwal, Nichol, Chu, and
  Chen]{ramesh2022hierarchical}
Aditya Ramesh, Prafulla Dhariwal, Alex Nichol, Casey Chu, and Mark Chen.
\newblock Hierarchical text-conditional image generation with clip latents.
\newblock \emph{arXiv preprint arXiv:2204.06125}, 2022.

\bibitem[Rombach et~al.(2022)Rombach, Blattmann, Lorenz, Esser, and
  Ommer]{rombach2022high}
Robin Rombach, Andreas Blattmann, Dominik Lorenz, Patrick Esser, and Bj{\"o}rn
  Ommer.
\newblock High-resolution image synthesis with latent diffusion models.
\newblock In \emph{CVPR}, pages 10684--10695, 2022.

\bibitem[Wang et~al.(2022)Wang, Guo, Hu, Chang, and Lyu]{wang2022gan}
Xin Wang, Hui Guo, Shu Hu, Ming-Ching Chang, and Siwei Lyu.
\newblock Gan-generated faces detection: A survey and new perspectives.
\newblock \emph{arXiv preprint arXiv:2202.07145}, 2022.

\bibitem[Verdoliva(2020)]{verdoliva2020media}
Luisa Verdoliva.
\newblock Media forensics and deepfakes: an overview.
\newblock \emph{IEEE Journal of Selected Topics in Signal Processing},
  14\penalty0 (5):\penalty0 910--932, 2020.

\bibitem[bel()]{bellingcat}
Bellingcat.
\newblock URL \url{https://www.bellingcat.com}.

\bibitem[Ho et~al.(2020)Ho, Jain, and Abbeel]{ho2020denoising}
Jonathan Ho, Ajay Jain, and Pieter Abbeel.
\newblock Denoising diffusion probabilistic models.
\newblock \emph{Advances in Neural Information Processing Systems},
  33:\penalty0 6840--6851, 2020.

\bibitem[Goodfellow et~al.()Goodfellow, Pouget-Abadie, Mirza, Xu, Warde-Farley,
  Ozair, Courville, and Bengio]{goodfellowgenerative}
Ian~J Goodfellow, Jean Pouget-Abadie, Mehdi Mirza, Bing Xu, David Warde-Farley,
  Sherjil Ozair, Aaron Courville, and Yoshua Bengio.
\newblock Generative adversarial networks.
\newblock In \emph{Advances in Neural Information Processing Systems}, pages
  2672--2680.

\bibitem[thi()]{this_city_does_not_exist}
This city does not exist.
\newblock URL \url{http://thiscitydoesnotexist.com/}.

\bibitem[dee()]{deepfake-satellite-images}
Dataset: deepfake-satellite-images.
\newblock URL \url{https://github.com/RijulGupta-DM/deepfake-satellite-images}.

\bibitem[Marín and Escalera(2021)]{rs13193984}
Javier Marín and Sergio Escalera.
\newblock Sssgan: Satellite style and structure generative adversarial
  networks.
\newblock \emph{Remote Sensing}, 13\penalty0 (19), 2021.
\newblock ISSN 2072-4292.
\newblock \doi{10.3390/rs13193984}.
\newblock URL \url{https://www.mdpi.com/2072-4292/13/19/3984}.

\bibitem[Zhao et~al.(2021)Zhao, Zhang, Xu, Yifan, and Deng]{Zhao_Deep}
Bo~Zhao, Shaozeng Zhang, Chunxue Xu, Sun Yifan, and Chengbin Deng.
\newblock Deep fake geography? when geospatial data encounter artificial
  intelligence.
\newblock 04 2021.
\newblock \doi{10.1080/15230406.2021.1910075}.

\bibitem[Zaytar and El~Amrani(2021)]{Zaytar_Satallite}
Mohamed Zaytar and Chaker El~Amrani.
\newblock Satellite image inpainting with deep generative adversarial neural
  networks.
\newblock \emph{IAES IJ-AI}, 10:\penalty0 121, 03 2021.
\newblock \doi{10.11591/ijai.v10.i1.pp121-130}.

\bibitem[Abady et~al.(2020)Abady, Barni, Garzelli, and Tondi]{abady2020gan}
L~Abady, M~Barni, A~Garzelli, and B~Tondi.
\newblock Gan generation of synthetic multispectral satellite images.
\newblock In \emph{Image and Signal Processing for Remote Sensing XXVI}, volume
  11533, pages 122--133. SPIE, 2020.

\bibitem[Abady et~al.(2022)Abady, Cannas, Bestagini, Tondi, Tubaro, and
  Barni]{abady2022overview}
Lydia Abady, Edoardo~Daniele Cannas, Paolo Bestagini, Benedetta Tondi, Stefano
  Tubaro, and Mauro Barni.
\newblock An overview on the generation and detection of synthetic and
  manipulated satellite images.
\newblock \emph{arXiv preprint arXiv:2209.08984}, 2022.

\bibitem[Corvi et~al.(2022)Corvi, Cozzolino, Zingarini, Poggi, Nagano, and
  Verdoliva]{corvi2022detection}
Riccardo Corvi, Davide Cozzolino, Giada Zingarini, Giovanni Poggi, Koki Nagano,
  and Luisa Verdoliva.
\newblock On the detection of synthetic images generated by diffusion models.
\newblock \emph{arXiv preprint arXiv:2211.00680}, 2022.

\bibitem[Ricker et~al.(2022)Ricker, Damm, Holz, and Fischer]{ricker2022towards}
Jonas Ricker, Simon Damm, Thorsten Holz, and Asja Fischer.
\newblock Towards the detection of diffusion model deepfakes.
\newblock \emph{arXiv preprint arXiv:2210.14571}, 2022.

\bibitem[Sha et~al.(2022)Sha, Li, Yu, and Zhang]{sha2022fake}
Zeyang Sha, Zheng Li, Ning Yu, and Yang Zhang.
\newblock De-fake: Detection and attribution of fake images generated by
  text-to-image diffusion models.
\newblock \emph{arXiv preprint arXiv:2210.06998}, 2022.

\bibitem[Lugmayr et~al.(2022)Lugmayr, Danelljan, Romero, Yu, Timofte, and
  Van~Gool]{lugmayr2022repaint}
Andreas Lugmayr, Martin Danelljan, Andres Romero, Fisher Yu, Radu Timofte, and
  Luc Van~Gool.
\newblock Repaint: Inpainting using denoising diffusion probabilistic models.
\newblock In \emph{CVPR}, pages 11461--11471, 2022.

\bibitem[Viquerat(2022)]{GitHubJviqueratbshapes}
Jonathan Viquerat.
\newblock {G}it{H}ub - jviquerat/bshapes: {A} 2{D} shape generator using
  {B}ezier curves.
\newblock \url{https://github.com/jviquerat/bshapes}, 2022.

\bibitem[Greig et~al.(1989)Greig, Porteous, and Seheult]{greig1989exact}
Dorothy~M Greig, Bruce~T Porteous, and Allan~H Seheult.
\newblock Exact maximum a posteriori estimation for binary images.
\newblock \emph{Journal of the Royal Statistical Society: Series B
  (Methodological)}, 51\penalty0 (2):\penalty0 271--279, 1989.

\bibitem[Radford et~al.(2021)Radford, Kim, Hallacy, Ramesh, Goh, Agarwal,
  Sastry, Askell, Mishkin, Clark, et~al.]{radford2021learning}
Alec Radford, Jong~Wook Kim, Chris Hallacy, Aditya Ramesh, Gabriel Goh,
  Sandhini Agarwal, Girish Sastry, Amanda Askell, Pamela Mishkin, Jack Clark,
  et~al.
\newblock Learning transferable visual models from natural language
  supervision.
\newblock In \emph{ICML}, pages 8748--8763. PMLR, 2021.

\bibitem[Gupta et~al.(2019)Gupta, Goodman, Patel, Hosfelt, Sajeev, Heim, Doshi,
  Lucas, Choset, and Gaston]{gupta2019creating}
Ritwik Gupta, Bryce Goodman, Nirav Patel, Ricky Hosfelt, Sandra Sajeev, Eric
  Heim, Jigar Doshi, Keane Lucas, Howie Choset, and Matthew Gaston.
\newblock Creating xbd: A dataset for assessing building damage from satellite
  imagery.
\newblock In \emph{CVPR workshops}, pages 10--17, 2019.

\bibitem[Wang et~al.(2020)Wang, Wang, Zhang, Owens, and Efros]{wang2020cnn}
Sheng-Yu Wang, Oliver Wang, Richard Zhang, Andrew Owens, and Alexei~A Efros.
\newblock Cnn-generated images are surprisingly easy to spot... for now.
\newblock In \emph{CVPR}, pages 8695--8704, 2020.

\bibitem[Mayer and Stamm(2020)]{mayer2020exposing}
Owen Mayer and Matthew~C Stamm.
\newblock Exposing fake images with forensic similarity graphs.
\newblock \emph{IEEE Journal of Selected Topics in Signal Processing},
  14\penalty0 (5):\penalty0 1049--1064, 2020.

\bibitem[Huh et~al.(2018)Huh, Liu, Owens, and Efros]{huh2018fighting}
Minyoung Huh, Andrew Liu, Andrew Owens, and Alexei~A Efros.
\newblock Fighting fake news: Image splice detection via learned
  self-consistency.
\newblock In \emph{ECCV}, pages 101--117, 2018.

\bibitem[Cozzolino and Verdoliva(2019)]{cozzolino2019noiseprint}
Davide Cozzolino and Luisa Verdoliva.
\newblock Noiseprint: A cnn-based camera model fingerprint.
\newblock \emph{IEEE Trans. on Information Forensics and Security},
  15:\penalty0 144--159, 2019.

\bibitem[Ng et~al.(2004)Ng, Chang, and Sun]{ng2004data}
Tian-Tsong Ng, Shih-Fu Chang, and Q~Sun.
\newblock A data set of authentic and spliced image blocks.
\newblock \emph{Columbia University, ADVENT Technical Report}, 4, 2004.

\bibitem[de~Carvalho et~al.(2013)de~Carvalho, Riess, Angelopoulou, Pedrini, and
  de~Rezende~Rocha]{Carvalho_dso1}
Tiago~José de~Carvalho, Christian Riess, Elli Angelopoulou, Hélio Pedrini,
  and Anderson de~Rezende~Rocha.
\newblock Exposing digital image forgeries by illumination color
  classification.
\newblock \emph{IEEE Trans. on Information Forensics and Security}, 8\penalty0
  (7):\penalty0 1182--1194, 2013.
\newblock \doi{10.1109/TIFS.2013.2265677}.

\end{thebibliography}

\onecolumn
\section{Additional Pristine Imagery}
                        
\begin{figure*}[hbt!]
    \centering
    
    \includegraphics[width=.8\textwidth]{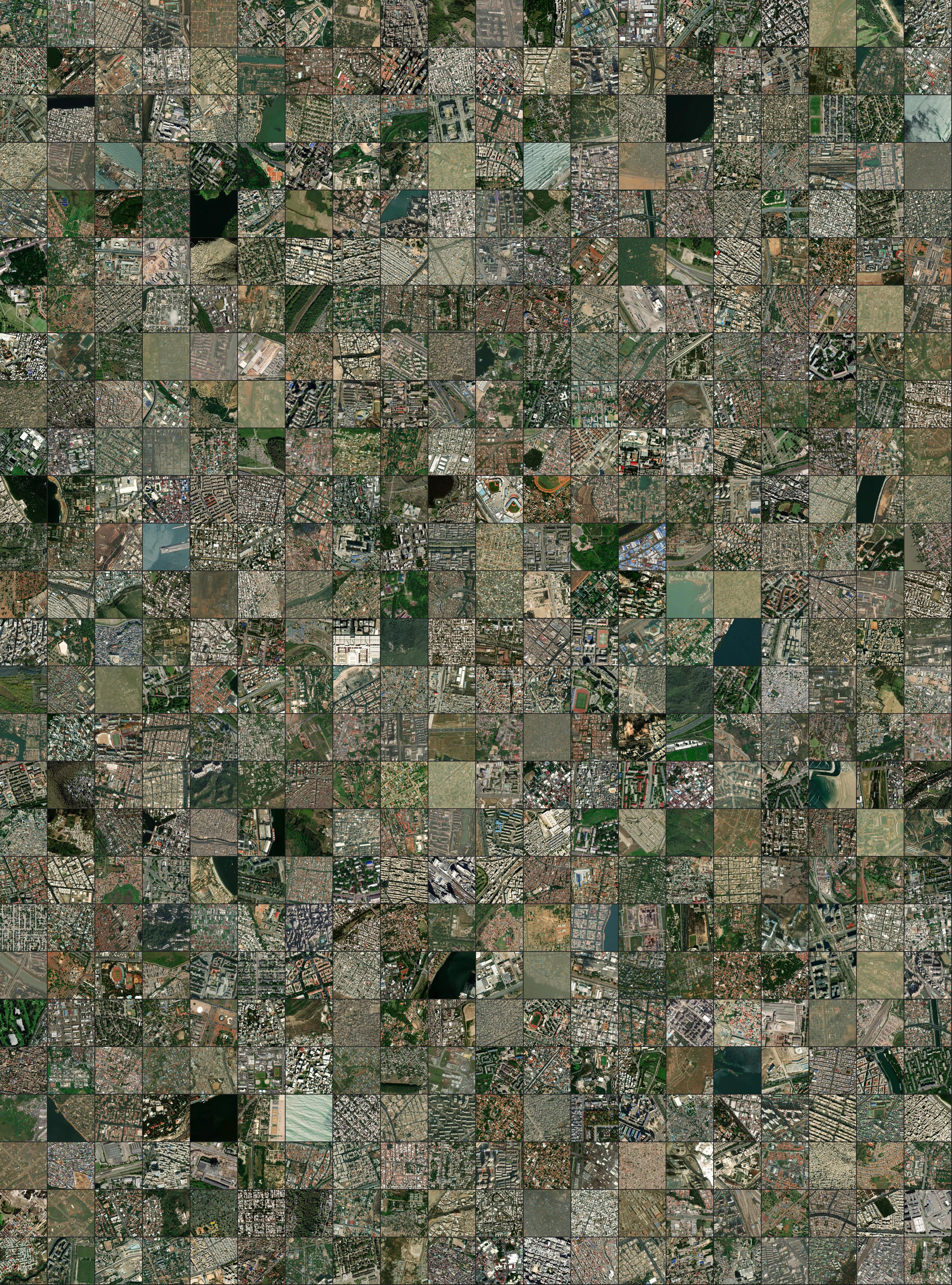}
    
    \caption{Collage of a representative sample of pristine images from cities around the world present in our dataset.}
    \label{fig:my_label}
\end{figure*}

\end{document}